\documentclass[journal]{IEEEtran}

\usepackage{cite}
\usepackage{amsmath,amssymb,amsfonts}
\usepackage{graphicx}
\usepackage{textcomp}

\usepackage{booktabs}
\usepackage{amsmath}
\usepackage{amssymb}
\usepackage{amsfonts}
\usepackage{graphicx}
\usepackage{amsthm}
\usepackage{xcolor}
\usepackage{mathtools}
\usepackage{float}
\usepackage{pgfplots}
\pgfplotsset{compat=1.7}
\usepackage[vlined, ruled, shortend, linesnumbered]{algorithm2e}

\usepackage[style=base]{caption}
\usepackage{subcaption}
\usepackage{hyperref}
\usepackage{tabstackengine}
\setstackEOL{\cr}

\usepackage{soul}

\newcommand\mydots{\hbox to 1em{.\hss.\hss.}}

\newlength\figureheight
\newlength\figurewidth
\SetKwBlock{kwDef}{Define:}{end;}
\SetKwBlock{kwGen}{Generate:}{end;}
\SetKwBlock{kwImp}{Implement:}{end;}
\SetKwBlock{kwCal}{Calculate:}{end;}
\SetKwBlock{kwInv}{Invoke:}{end;}
\SetKwBlock{kwCbk}{Callback:}{end;}

\begin{document}

\title{
    Secure Encoded Instruction Graphs for \\ End-to-End 
Data Validation in Autonomous Robots}

\author{Jorge Peña Queralta,
        Li Qingqing,
        Eduardo Castelló Ferrer
        and~Tomi Westerlund
\thanks{Jorge Peña Queralta, Li Qingqing and Tomi Westerlund are with the \href{https://tiers.utu.fi}{Turku Intelligent Embedded and Robotic Systems (TIERS) Lab, University of Turku, Turku, Finland}, e-mails: \{jopequ, qingqli, tovewe\}@utu.fi. Eduardo Castelló Ferrer is with the \href{https://connection.mit.edu/}{MIT Connection Science} and \href{https://www.media.mit.edu/}{MIT Media Lab}, Massachusetts Institute of Technology, Cambridge, USA. e-mail: ecstll@mit.edu}}

\markboth{IEEE INTERNET OF THINGS JOURNAL}{Peña Queralta \MakeLowercase{\textit{et al.}}: Secure Encoded Instruction Graphs for End-to-End Data Validation in Autonomous Robots}


\maketitle

\begin{abstract}
    As autonomous robots are becoming more widespread, more attention is being paid to the security of robotic operation. Autonomous robots can be seen as cyber-physical systems: they can operate in virtual, physical, and human realms. Therefore, securing the operations of autonomous robots requires not only securing their data (e.g., sensor inputs and mission instructions) but securing their interactions with their environment. There is currently a deficiency of methods that would allow robots to securely ensure their sensors and actuators are operating correctly without external feedback. This paper introduces an encoding method and end-to-end validation framework for the missions of autonomous robots. In particular, we present a proof of concept of a map encoding method, which allows robots to navigate realistic environments and validate operational instructions with almost zero {\it a priori} knowledge. We demonstrate our framework using two different encoded maps in experiments with simulated and real robots. Our encoded maps have the same advantages as typical landmark-based navigation, but with the added benefit of cryptographic hashes that enable end-to-end information validation. Our method is applicable to any aspect of robotic operation in which there is a predefined set of actions or instructions given to the robot.
\end{abstract}

\IEEEpeerreviewmaketitle

\section{Introduction}

\begin{figure}
    \centering
    \includegraphics[width=0.48\textwidth]{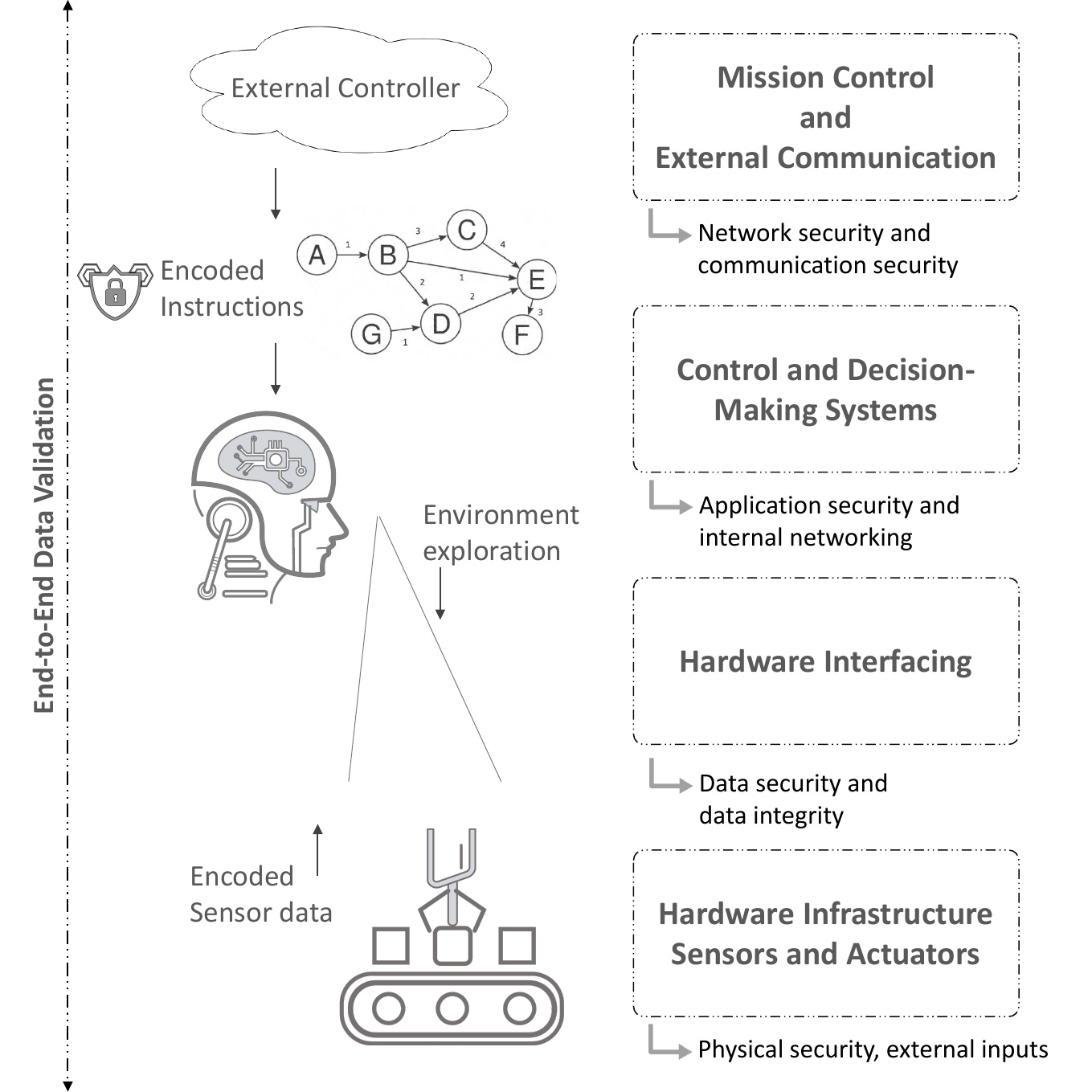}
    \caption{Classification of data acquisition and analysis processes in autonomous robots and matching security layers.}
    \label{fig:concept}
\end{figure}

As robots become increasingly common across many domains and application areas, more attention is being paid to the safety and security aspects of robotic operation~\cite{kirschgens2018robot}. The differentiation between safety and security is often ambiguous, with the term \textit{safety} typically being used to refer to human--robot interaction~\cite{bragancca2019brief} or to the protection of the robot from physical damage~\cite{huang2014rosrv}. However, what is typically overlooked is that safe operation of an autonomous robot is inherently tied to having tight security of the data involved, including sensor data and data defining mission instructions. Fig.~\ref{fig:concept} shows a classification of stages in which information is either collected or processed by an autonomous robot. The classification in Fig.~\ref{fig:concept} extends the cyberattacks categorization defined in~\cite{clark2017cybersecurity}. It also models the software processes as a network~\cite{akhunzada2015securing}, which corresponds with many robotic frameworks, such as the Robot Operating System (ROS)~\cite{rivera2019ros}. In this classification in Fig.~\ref{fig:concept}, the acquired sensor data needs to be both secured and validated, which is essential from a cybersecurity point of view and represents an unresolved challenge. For secure operation of autonomous robots, it is essential to be able to validate the data being shared among subsystems and external systems (e.g., a controller or other robots) and the data defining or characterizing the way the robot interacts with its environment.

A relevant precedent in securing multi-robot cooperation was introduced by Castell\'o Ferrer et al. in~\cite{ferrer2019secure}, in which the authors leveraged Merkle trees to improve the security and secrecy of swarm robotics missions. The main novelty of their work is the introduction of a framework for validating data in robots without relying on the data itself, by encoding mission instructions in Merkle trees. Merkle trees are cryptographic structures that enable validation of data through cryptographic proofs that do not involve the data itself.

We aim to extend the previous work to build a more general framework: rather than encoding one set of mission instructions~\cite{ferrer2019secure}, we encode all possible ways in which a mission can be completed. In~\cite{ferrer2019secure}, one of the main research questions is whether it is possible to provide the \textit{blueprint} of a robotic mission without describing the mission itself.
In this work, we build upon the separation of data verification from the data itself, as introduced in~\cite{ferrer2019secure}, and explore new implicit ways for defining complex robotic workflows. 
In~\cite{ferrer2019secure}, the robots perform predefined actions when they are able to reproduce some predefined encoded sensor data. 
In this work, we instead utilize a graph structure with connections between the possible mission instructions or states, in which paths through the graph encode all possible mission flows. 

In this manuscript, we first describe the proposed framework and several possible applications, including human-robot interaction and collaborative robots. We then provide a proof of concept in an example navigation mission. The proof of concept demonstrates that our framework has minimal impact on robot behavior and robustness, even when the full mission is encoded. Our framework is therefore viable for real applications, and opens the door to more secure and safe deployment of autonomous robots.

We summarize the current research gap in robust data validation schemes for autonomous or semi-autonomous robots with the following key unanswered research questions.
\begin{enumerate}
    \item Is it possible for a robot to safely and securely interact with its environment, operators, and other robots without {\it a priori} knowledge of the mission?
    \item Can encoded information that reveals no explicit mission instructions maintain the performance of unencoded robotic workflows?
    \item Can encoded mission instructions be used to simultaneously validate the integrity of robot actions, the progress of the mission, the sensor data, and the operations of actuators?
\end{enumerate}

The first question applies to a wide variety of situations. For instance, consider whether a robot in a factory can be given assembly instructions that it cannot understand until reaching the respective step in the assembly process, or whether a mobile robot can be given a map that it cannot understand until it is navigating the respective location in its environment. An even more complex situation occurs when a robot can interact with a human or another robot in a certain way only when a series of conditions are met. The latter two questions deal with practical considerations: whether encoded instructions can be adapted to standard robotic algorithms and workflows and whether robots can use encoded instructions to simultaneously validate all processes involved in a mission. In short, we are asking whether a single framework can provide end-to-end data validation and provide encoded instructions that enable a robot to securely and safely interact with its environment.

We consider the validation of a robot's operation from end to end: validating sensor data, the correct operation of actuators, mission instructions, and information received from an external controller. We conduct proof-of-concept experiments in a navigation mission in which limited {\it a priori} information about the environment is available to the mission controller or human operator. In this scheme, an operator generates a set of encoded instructions by hashing the description of a set of landmarks in the environment that serve as navigation waypoints. The encoded set of landmarks is given to the autonomous robots, which use them to navigate a realistic environment without any other {\it a priori} knowledge. The encoded landmarks are nodes in a navigation map that is given to the robots in the form of a graph. The graph edges encode information about how to navigate between consecutive landmarks. Because all information is encoded, we minimize the amount of raw data exposed to the robots {\it a priori}.

The main contributions of this work are:
\begin{enumerate}
    \item the definition of an end-to-end validation framework for autonomous robots based on encoded instruction graphs;
    \item the definition and validation of a novel approach to use cryptographic hashes to encode a navigation graph that maps environment features;  and
    \item the definition and validation of methods that allow robots to follow encoded mission instructions while validating their sensor data without external feedback.
\end{enumerate}

The remainder of this paper is organized as follows. In Section~II, we overview previous works that address security issues in robotic systems. Section~III then introduces the proposed framework, describes different use cases, and discusses the potentials and limitations of our framework. Section~IV presents the implementation details of a proof of concept for robotic navigation, using an encoded navigation graph. Then, Section~V describes the methodology we have followed in our simulations and experiments and Section~VI presents the experimental results. Finally, Section~VII concludes the work and outlines future research directions.

\begin{figure*}
    \centering
    \includegraphics[width=\textwidth]{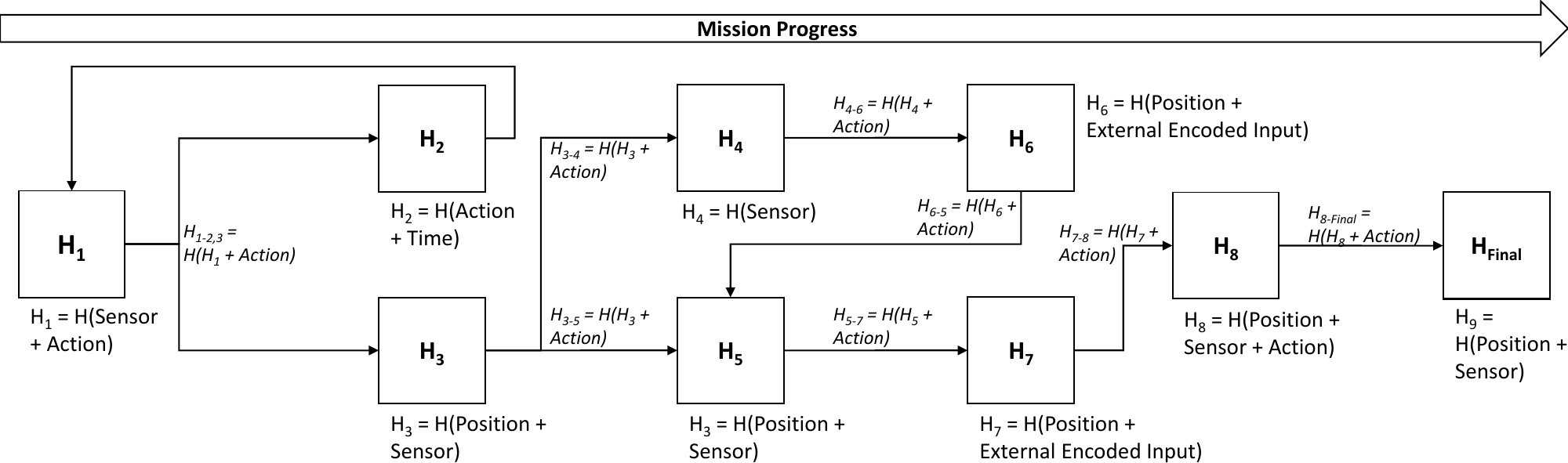}
    \caption{Sample encoded instruction graph, with hashes that are defined from different environmental information, or combinations of sensor data, localization data, and one or more of the available actions. The mission can be completed successfully only if the hashes are decoded sequentially according to one of the paths encoded in the graph. Note that there is not necessarily a single way of progressing in the mission.}
    \vspace{-1em}
    \label{fig:sample}
\end{figure*}

\section{Background}

There is a growing interest in securing robotic systems, partly due to to the increasing connectivity with which robots are being equipped. There are many data exchange modalities with new attack vectors, for instance in remote control commands~\cite{liu2005ibotguard}, telemetry~\cite{rodday2016exploring}, offloading computation~\cite{wan2016cloud, qingqing2019odometry}, robot-to-robot communication~\cite{ferrer2019secure, ferrer2021leaders}, and human--robot interaction~\cite{miller2018social}.

Existing research efforts that study cybersecurity issues in robotics generally consider exclusively virtual or data aspects. In~\cite{clark2017cybersecurity}, Clark et al. review and discuss the main security threats to robotic systems, including spoofing sensor data, denial of service attacks, malicious code injection, and signal interference. However, this review only considers the cybernetic point of view, not explicitly the physical dimensions in which robots operate. 
Some works have considered security in the context of sensor data and navigation missions. For instance, in an early work in this direction, Py et al.~\cite{py2004dependable} introduced an execution control framework for an autonomous robot to analyze the data it obtained through its behaviors using a \textit{state checker}. In a more recent work, Tang et al.~\cite{tang2018event} ensured the convergence of sensor data under denial of service attacks. Similarly, in~\cite{tiku2019overcoming}, Tiku et al. introduced a methodology for overcoming security vulnerabilities in a deep learning localization method by making use of adversarial training samples.
In distributed and multi-robot systems, most efforts focus on the analysis and mitigation of security issues from a networking perspective~\cite{rahman2019evaluation}.
All of these approaches take the point of view of data security in information systems and do not explicitly involve the cyber-physical nature of autonomous robots and their interaction with their environment, which is the objective of this paper. 
The interaction of a robot with its environment presents new crucial issues that cannot be addressed with existing risk mitigation techniques from the cybersecurity domain.

There have been a few studies on data validation for autonomous robots. Legashev et al.~\cite{legashev201monitoring} define a generic framework from a legislative point of view, for monitoring, certification, and validation of the operation of autonomous robots, using periodic telemetry data obtained from autonomous vehicles. 
The framework developed by Legashev et al. validates a robot's operation but not the data itself. Validation of the data in this work can only be done through statistical analysis and detection of statistical abnormalities. Data integrity was the subject of one study by Yousef's et al.~\cite{yousef2018analyzing} on cyber-physical threats to robotic platforms.

In general, we see a research gap in terms of addressing the physical dimension of security issues in robotic operation. This becomes even more evident when considering widely used robotic frameworks, such as Robot Operating System (ROS), which has become a standard across both industry and academia. Multiple researchers have studied the security flaws of ROS~\cite{demarinis2019scanning}, and proposed approaches to address these issues~\cite{amini2018cryptoros}. Many of these are also being mitigated in the newest version, ROS~2~\cite{kim2018security}. However, these efforts are again mostly directed at securing ROS as a distributed and networked system, not at securing robots' interactions with their environment. While it is highly important to provide security for data flow, it is also important to close the gap in securing and validating the way robots receive instructions and interact with their environment.

In this work, we focus on a framework for validating data integrity. Other types of cyberattacks such as denial of service attacks---in which the communication channels are congested---are not considered. However, it is worth noting that our proposed approach can provide some benefits even in the types of cyberattacks we do not directly consider. For instance, while the communication channel used to transmit encoded commands to the robot might be known to an attacker, the sensor data or inputs triggering the different actions will still be unknown to an attacker, as they are unknown even to the robot itself. As another example, the channel utilized to trigger a robot's actions might be disguised within the encoded instruction graph sent to the robot before the mission starts. In these ways, our framework can provide partial mitigation for a broad range of possible cyberattacks.

\begin{figure*}
    \centering
    \includegraphics[width=0.96\textwidth]{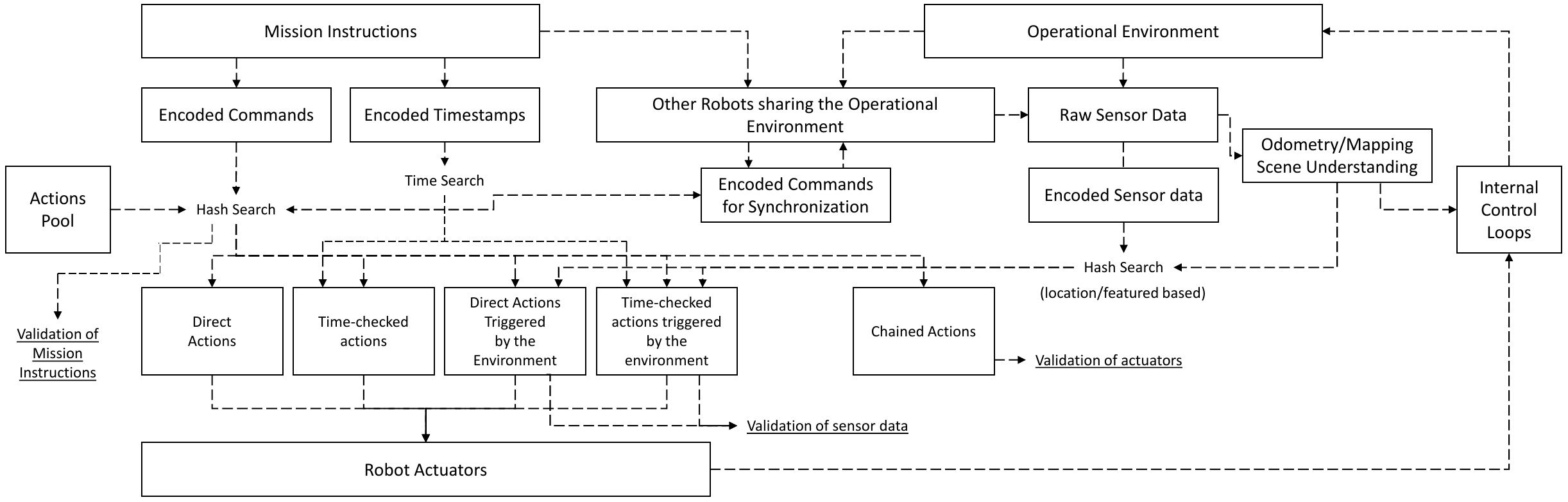}
    \caption{Different validation modalities and data flows. The same approach can be used for individual mission instructions, event-based commands, chained instructions, or multi-robot communication.} 
    \vspace{-1em}
    \label{fig:actions}
\end{figure*}

\section{Encoded Instruction Graphs Framework}

This section presents the framework and how it can generally be adapted to different application scenarios, before focusing in the remainder of the manuscript on how it can be applied to a specific navigation use case.

\subsection{Encoding Robotic Instructions}

We extend the instruction encryption ideas from~\cite{ferrer2019secure}, in which mission instructions are given to a robot by encoding combinations of sensor inputs and robot actions. We do not explicitly consider multi-robot cooperation at this point. We describe robust options for encrypting and decrypting mission instructions and evaluate the performance degradation that is inherent to the addition of data encryption to robotic operation. 

In order to encrypt a robotic operation, the first step is to encode a set of actions and features from sensor data along with at least one more variable that enables a \textit{hash search}. A hash search is a trial-and-error process in which a robot does not necessarily reproduce a specific hash but instead can try multiple hashes until finding a match. The additional variable that enables the hash search can for instance be a spatial or temporal component. The second step is to define a series of encoded actions and encoded states based on a combination of variables (e.g., position, time, sensor data, or other external inputs). By encoding both states and actions, we can wrap the set of encoded information into a graph structure. The encoded information in an edge of the graph gives the robot information about how to arrive to the respective state or process. We call this an encoded instructions graph. A sample encoded instructions graph is shown in Fig.~\ref{fig:sample}. In this graph, the initial instruction is encoded into a hash $H_1$, which the robot is able to decode by combining a predefined action (e.g., movement in one specific direction) with predefined sensor data (e.g., visual or geometric features extracted from the environment). Most of the nodes in the graph represent states. In the example shown in Fig.~\ref{fig:sample}, most of the nodes are defined {\it a priori} by the combination of a specific position with specific sensor data (associated to an environment feature, e.g., a predefined landmark). 

The information that nodes can encode is not restrictive, as it can be any action or state. The edges, however, need to encode information that enables the robot to transit between the respective nodes. Therefore, an edge needs to encode one of the robot's possible actions or an external input, such as a message from a controller or another robot. An edge can also be empty, e.g., if a robot at the previous node already has enough information to decode the next node, without any intermediate step. In the example in Fig.~\ref{fig:sample}, the first node triggers an action by the robot when its acquired sensor data reproduces the respective hash in the graph. The robot then proceeds to one of two different states in which it must to be able to reproduce both a specific position and a specific sensor reading. The encoded information in the second node (e.g., $H_2$) can be decoded after a certain period of time. The robot can be forced to go back to the previous state if the sensor data encoded in $H_3$ cannot be acquired in time, which can for example be used as a failsafe. In practice, the robot is not aware of the type of information encoded in each hash, and must therefore perform a continuous trial-and-error process to reproduce the hashes in the graph, by any means for which it has been preprogrammed. In our experiments, we show that this trial-and-error process has a mostly negligible impact on robot behavior, compared to the computational cost of extracting features or processing sensor data. However, the process of deciding how to encode the instructions is not trivial, as they must be reproducible, but also must be concise enough to avoid data mismatches.

The graph depicted in Fig.~\ref{fig:sample} is directed. In, for example, a mission with only one possible solution in which each step is followed by only one other step, the graph simply represents a linear sequence. In more complex missions, the graph will often remain acyclic, with multiple possible paths to complete the mission, but without possibility to repeat a step once it has been completed. For instance, in a manufacturing process, the order in which a set of parts are moved to a working bench might not matter, yet every part must be moved exactly once. The graph, nonetheless, can also contain cycles. For instance, in a reconnaissance mission in which a robot has to navigate an area but without a specific path, the graph would include an interconnected set of steps (i.e., a cycle). A robot's higher mission control should understand these situations and provide a planning strategy that is not directly sent by the mission controller (e.g., path planning on the encrypted graph). An example of this is shown in Sec.~\ref{sec:Results}.

\subsection{Validation Modalities}
The proposed approach can be applied to many types of scenarios, because the encoded information in the graph is not necessarily limited to a set of predefined actions and features extracted from sensor data. Other external inputs can also be included, such as variables defining the state of the robot or timing constraints. The addition of these types of variables brings the possibility to enable secure and secret multi-robot cooperation as well as new ways of defining the conditions under which human-robot interaction can happen. The node inputs are all encoded, so the information exchanged between robots or utilized as external signals triggering robot actions is only usable when combined with other data and is totally meaningless for an external agent. Therefore, if the data is spoofed or a third party gains access to this communication medium, no meaningful data is actually compromised. 

Different possibilities for encoding and decoding instructions are shown in Fig.~\ref{fig:actions}. In the figure, we show the different approaches to encoding a robot's interaction with its environment. For example, these options include simply encoding a predefined action (direct actions); combining actions with a timestamp indicating when they should be carried out (time-checked actions); combining actions with specific sensor data obtained from the environment (direct actions triggered by the environment); or a combination of all the previous options. These actions can then be chained, such that the previous hash or action is included in the next codified node in the graph. These approaches are described in more detail and put into context in the following subsections.

\subsubsection{Independent Validation}
The simplest approach is to encode mission instructions individually and independently. In this case, an encoded instruction set can be sent to the robot, similar to the approach followed in~\cite{ferrer2019secure}. However, this set of instructions does not define a graph structure and does not represent the main interest of this paper. 

An additional layer of security can be added by introducing time or spatial constraints. Time constraints (e.g., hashing a timestamp together with the data of interest) provide an extra layer of security against attacks that could spoof the encoded data and reproduce it at a different time. Similarly, spatial constraints can be added by including the robot's location in the hash. This, however, only prevents the replication of the robot's behavior in other locations.

\subsubsection{Iterative Validation}
An iterative validation happens when a robot is able to validate its own actions. This modality is studied in the next sections with the introduction of an encoded navigation graph.

\begin{figure}
    \centering
    \includegraphics[width=0.48\textwidth]{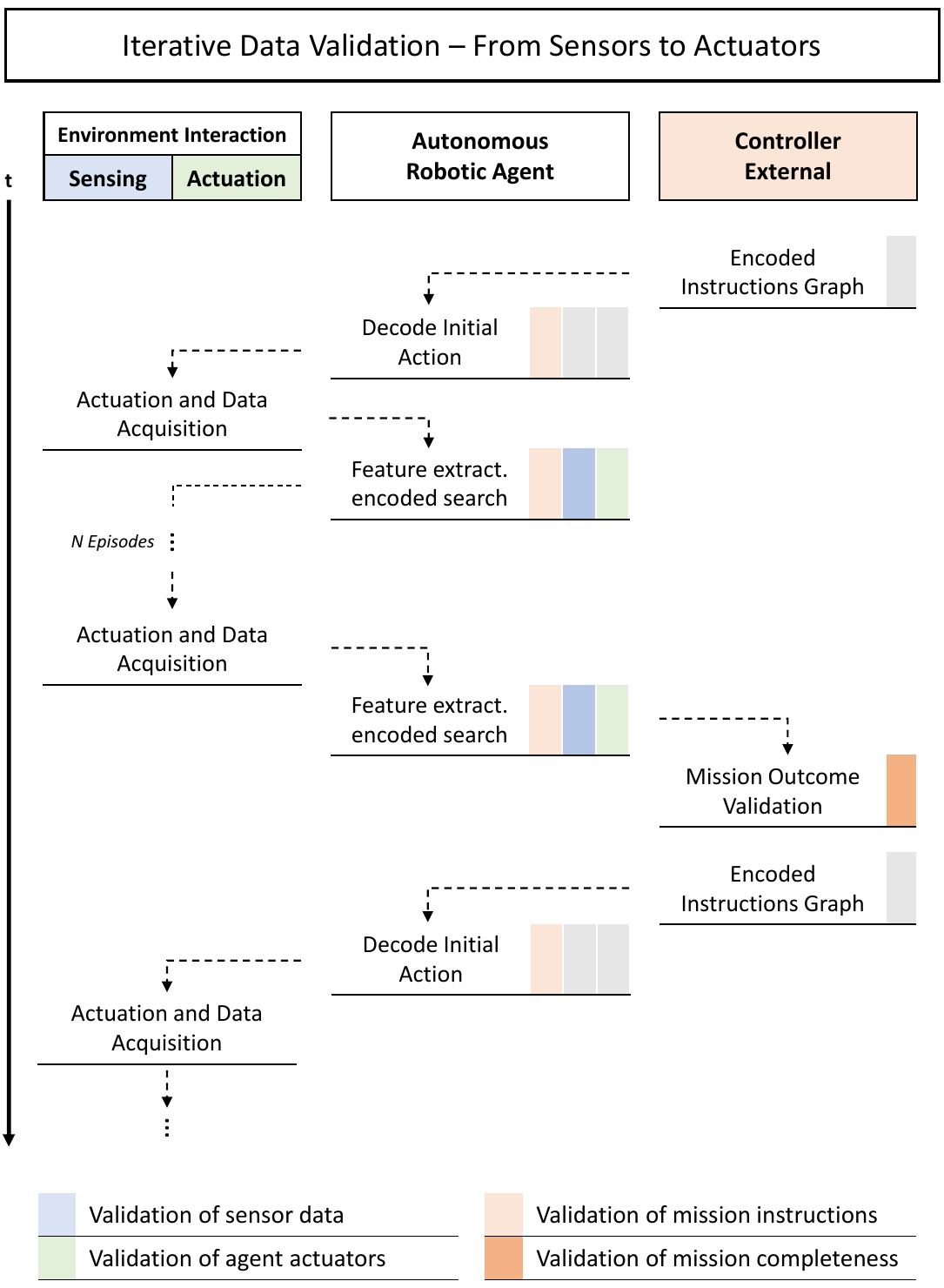}
    \caption{Illustration of an iterative validation process.}%
    \vspace{-1em}
    \label{fig:iterative}
\end{figure}

Encoded instruction graphs defining an iterative validation process can contain different types of encoded data in their nodes and edges. For instance, sensor data can be encoded in the graph nodes, which serve the purpose of process validation. Additionally, other information can be encoded, such as positional information or time information, which can be used in the control loops of the robot itself. Then, the actual instruction for the robot to move towards the next step is encoded in the edge of the graph, which encodes both the data in the current node being validated and the action or actions that will enable the robot to decode the next node. An illustration of this process is shown in Fig.~\ref{fig:iterative}. In the figure, we show how, at each step, instructions can be validated simply by being able to reproduce the corresponding hash. Moreover, as actions accumulate leading to new reproducible states (defined through chained hashes in the encoded instruction graph), we are able to iteratively validate the actions confirming their output with an expected outcome. 

\subsubsection{Multi-Robot Simultaneous and Mutual Validation}
\begin{figure}
    \centering
    \includegraphics[width=0.48\textwidth]{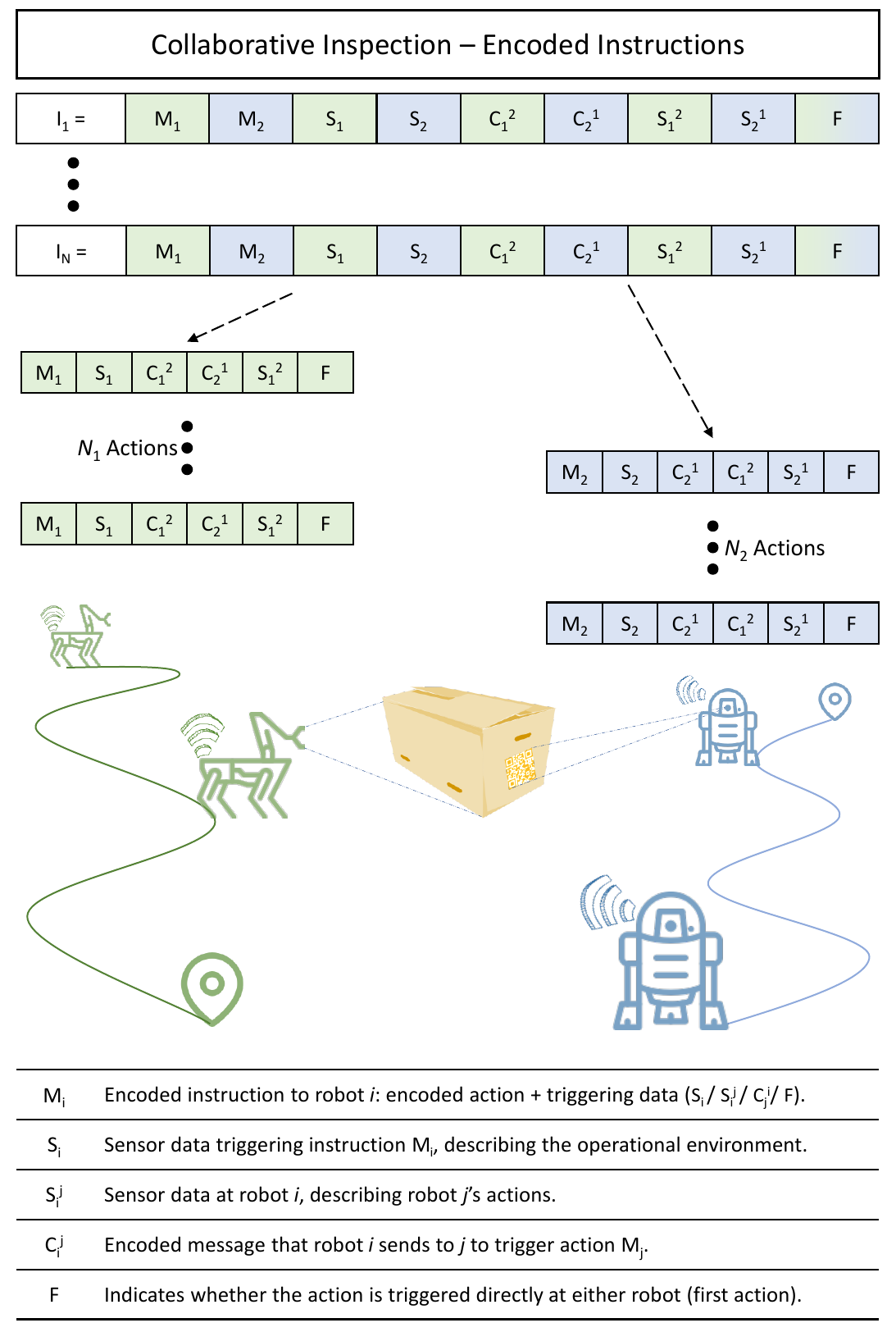}
    \caption{Illustration of a collaborative inspection process where robots only have partial instructions.}
    \label{fig:collaborative}
\end{figure}

As mentioned earlier, one important scenario to which this validation framework can be applied is multi-robot cooperative missions. Complementing the ideas proposed in~\cite{ferrer2019secure}, we are now also able to break down a mission into two parts that can be given to two different robots. An example of this is shown in Fig.~\ref{fig:collaborative}, which illustrates a collaborative inspection process. In the example, encoded instructions are defined by combining a set of different signals or parameters. First, a set of sensor data is set to trigger an action from features extracted by the robot from. Now that multiple robots share the same environment, we can also differentiate between hashes obtained from sensing the environment and those obtained from sensing the behavior of other robots. Second, the robots can also exchange messages in order to trigger each other's actions. These messages do not hold any valuable data to the sender but only to the receiver as part of a hash decoding process. The messages can be predefined and based on the robot state, or generated as a function of the features sensed in the environment.

\begin{figure*}
    \centering
    \begin{minipage}{0.51\textwidth}
        \centering
        \includegraphics[width=\textwidth]{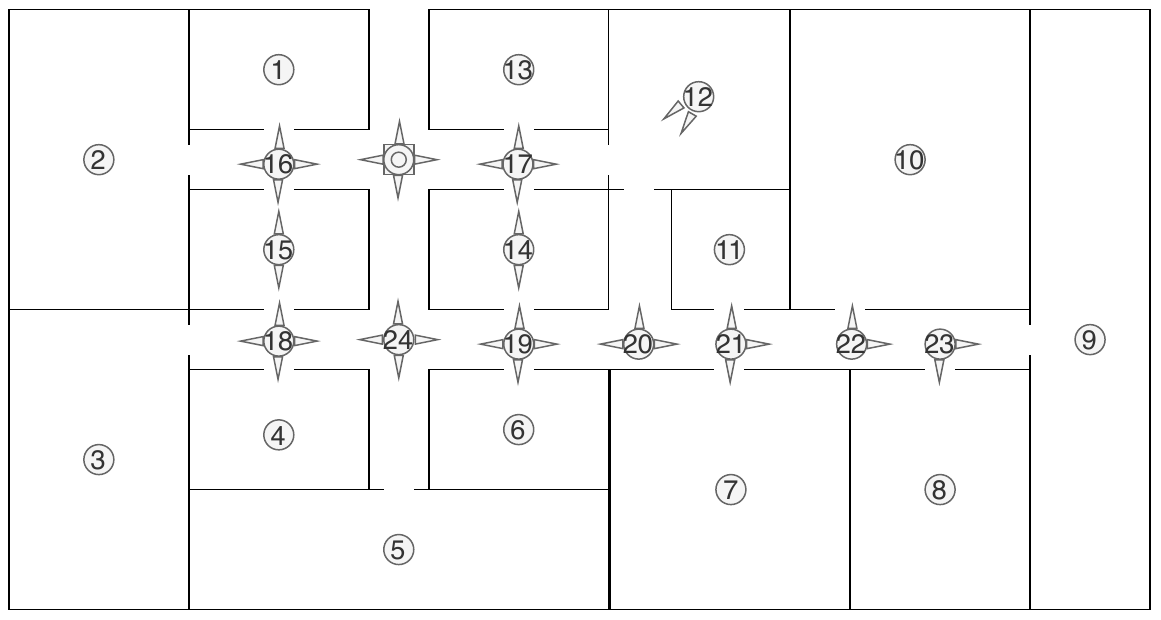} \\
        \small{(a)}
    \end{minipage}
    \begin{minipage}{0.48\textwidth}
        \centering
        \includegraphics[width=\textwidth]{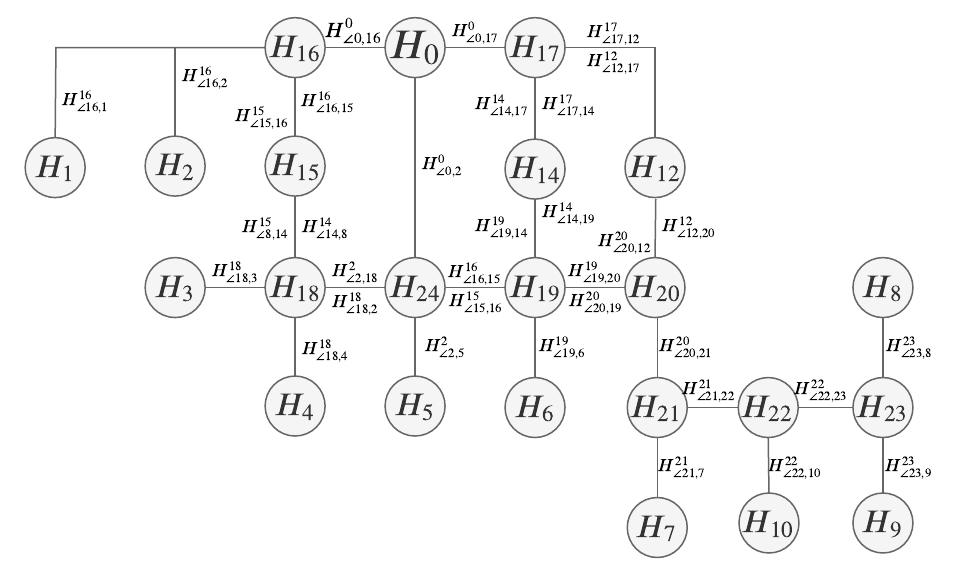} \\
        \small{(b)}
    \end{minipage}
    
    $ $ \\[+2pt]
    
    \begin{minipage}{0.45\textwidth}
        \centering
        $ $ \\[+8pt]
        \footnotesize
        \begin{tabular}{ll}
            \toprule
            \textbf{ID} & \textbf{$H_N$} (256 bits) \\
            \midrule
            0 & $H_0=$ \scriptsize{5341dfa945ca9e52334 \dots 8446048} \\
            1 & $H_0=$ \scriptsize{a6340c2ed22ff55a475 \dots c3e92b4}  \\
            \vdots & \\
            24 & $H_{24}=$ \scriptsize{2522cfaa21faaa45a \dots d9de50a} \\
            \bottomrule
        \end{tabular}
        $ $ \\[+6pt]
        \small{(c)}
    \end{minipage}
     \begin{minipage}{0.54\textwidth}
        \scriptsize
        \begin{equation*}
        \setlength\arraycolsep{3pt}
        \begin{pmatrix}
            0 & 0 & 0 & \mydots &  0 & H^{0}_{\angle0,16} & H^{0}_{\angle0,17} & \mydots & H^{0}_{\angle0,24} \\
            \vdots & \vdots & \vdots &  & \vdots  & \vdots & \vdots & & \vdots \\
            H^{16}_{\angle16,0} & H^{16}_{\angle16,1} & H^{16}_{\angle16,2} & \mydots & H^{15}_{\angle16,15} & 0 & 0 &  \mydots & 0 \\
            \vdots & \vdots  & \vdots &   & \vdots  & \vdots & \vdots &  & \vdots \\
            0 & 0 & 0 & \mydots & \mydots & 0 & 0 & \mydots & H^{24}_{\angle24,0} \\
        \end{pmatrix}
        \end{equation*}
        \centering
        $ $ \\
        \small{(d)}
    \end{minipage}
    \caption{Encoded navigation graph construction process: (a) shows a sample floorplan and (b) the corresponding navigation graph. Subfigures (c) and (d) contain the information given to robots, a list of hashes and an adjacency matrix with the edge hashes to aid navigation between landmarks, respectively. The landmark hashes are calculated based on the position $(x,y,z)$ and the landmark type ($LT$): $H_i = H(LT_i + x_i + y_i + z_i)$, where $H$ is the hashing function and $+$ means concatenation. The edge hashes are $H^{i}_{\angle i,j} = H(LT_i + x_i + y_i + z_i + \angle i,j)$, where $\angle i,j$ represents the navigation direction from $i$ to $j$.}
    \label{fig:sample_map}
\end{figure*}

\section{Use Case Encoded Navigation Graph}
\label{section.EncodedNavigationGraph}

One of the most fundamental ways in which a robot interacts with its environment is through navigation. Maps have long been utilized for autonomous navigation and exploration in mobile robots to increase the robustness of long-term autonomous operation~\cite{hilnbrand2019roadwaygeometries, sobreira2019mapmatching}. Maps and landmarks provide robots means for localization in a known reference frame, while enabling the calibration and adjustment of on-board odometry and localization algorithms.

Landmark-based navigation has been successfully implemented in various mobile robots with quick response (QR) codes~\cite{andersen2013qrcalibration, zhang2015qrnavigation, nazemzadeh2017qrfusion} or other identifiable images\cite{zamir2010accurate, sattler2016efficient, thoma2019mapping}, wireless sensor networks~\cite{cheng2011wsn}, ultra-wideband (UWB) markers~\cite{almansa2020autocalibration, shule2020uwbbased, song2019uwb}, IMU fusion~\cite{nazemzadeh2017qrfusion}, or topological maps for infrastructure-free navigation~\cite{gadd2015warehouse}. When utilizing landmarks that are already encoding certain information, such as QR codes or other text representations, additional information can be embedded into the landmarks. In an industrial scenario, this can be utilized to provide further instructions for robots~\cite{qin2017sorting}. 

In order to analyze the viability of the proposed research, we decided to investigate the problem of robot navigation since it is an easy way to make the robot interact with its environment. Along those lines, we focus the research questions to more concrete considerations regarding the navigation of autonomous robots:

\begin{enumerate}
    \item Is it possible to provide a description of the environment (e.g., a map or a set of landmarks and how to travel between them) to an autonomous robot in such a way that the robot is unable to understand the map until it starts navigating, and such that it can only decode the information in that map if a series of conditions on how it sees its environment are met? 
    \item Is there a way of defining navigation instructions for an autonomous robot such that any modification of those instructions automatically renders them unusable ensuring that if wrong sensor data is fed to the robot's controller, the instructions cannot be followed?
\end{enumerate}

\subsection{Encoded Graph Definition}
Rather than modeling a map of the exploration area and utilize it for navigation, we utilize a landmark-based navigation graph that encodes the position of the different landmarks and the navigable directions between them. In this graph, each vertex represents one encoded position in the map, and each edge represents a straight or unique path between two positions. By unique we mean a path that might not be straight but such that the robot can realistically follow. A sample map and the corresponding encoded navigation graph are illustrated in Fig.~\ref{fig:sample_map}. In this and latter sections, we utilize the following notation: a graph is an ordered pair $\mathcal{G}=(V, E)$, where $V$ represents a set of vertices, and $E$ represents a set of edges associated with two distinct vertices, i.e., a set of tuples $\{(V_i,V_j) \:|\: V_i,V_j \in V\}$. We consider a directed graph, were the order of these tuples matters.

The most straightforward approach to landmark encoding is to define the hash of a position given its coordinates $\vec{r}\in\mathbb{R}^3$. Thus we would define $H_i=H(\vec{r}_i)$. In order to ensure that hashes will be reproducible, the coordinates need to be given in a coarse grid with a resolution that is dependent on the accuracy of the robots' onboard odometry. 

If the environment is accessible {\it a priori}, elements such as QR codes or Bluetooth/UWB beacons can be placed to facilitate the localization of robots when they are nearby. The QR codes contain hashed data and can encode additional information, for example, instructions for a robot to operate in a given room or area. An alternative approach is to utilize the environment geometry and topology. The coordinates of the features can still be utilized to define their hash without using a predefined grid. Rather than having a robot utilizing its own or near position to calculate the hash, it can calculate it based on the coordinates of a position that depends on the robot's current local environment. 

\subsection{Deployment and Navigation}
We assume that the initial location where a robot is deployed is either known in an absolute reference frame, or utilized as a common reference in the robots' local coordinate system. If only local references are utilized, these must have a common orientation. This initial location is encoded with a hash but is also known to robots.

In the encoded navigation graph, each edge in the graph is given two hashes, as the robots might reach these from different directions. Therefore, the adjacency matrix containing the edge hashes shown in Fig.~\ref{fig:sample_map} is not symmetric. Only minimal information about the local environment required for navigation purposes needs to be stored at the robots. Odometry-only navigation, when possible, is preferred to map-based navigation to minimize the amount of raw information that robots store.

Global locations and relative positions between features as well as directions between them are encoded in a way that can be matched by robots on the basis of trying multiple possible directions until finding one that produces the corresponding hash. The edge hashes are calculated on a trial-and-error basis, and thus they can be defined with an arbitrary division of the $[0, 2\pi]$ interval. However, the granularity of such division must consider the inherent computational overhead. Furthermore, not all the navigable directions are necessarily selected, and therefore the real topology of the environment can be, to some extent, hidden. In addition, multiple features can be selected within a single room or small area, but even if all are detectable at the same time, a fully connected subgraph does not need to be generated within the navigation graph. In general terms, there is a trade-off between the number of paths in the navigation graph between two features, and the robustness of the navigation. In general, the more paths between two events, the higher the chance that the robot is able to reproduce a certain subset of hashes.

\subsection{Landmark-based Localization}
The accuracy of the feature's position directly affects the error tolerance for the odometry method utilized when no landmarks are detected. To cope with this issue, the odometry error can be estimated and then taken into account to calculate the hashes from the position of landmarks. This is achieved by following a trial-and-error approach within a certain spatial area around the landmark. The number of trials that a robot needs to perform depends on the accuracy of the odometry method as well as the granularity of the grid utilized to define the position of the landmarks. Additionally, the possibility of the robot identifying a wrong landmark that is nearby must be taken into account.

\section{Navigation Graphs: Methodology}

To test the feasibility of the encoded navigation approach presented in this paper, we conducted a series of simulations and real-robot experiments. In these, we analyzed the computational overhead as well as the performance impact of utilizing an encoded navigation graph. In all cases, we made the assumption that the environment is known to the mission controller. We devised two types of application scenarios in which we test the proposed framework.

First, we considered an environment where only robots are present. In this case, we simulated the interior of a building with empty rooms. For this environment, we encoded geometric features in the navigation graph such as doorways, corners and rooms. In our simulations, we considered an environment with no dynamic obstacles and a known geometry. This environment is comparable to modern logistic warehouses where only autonomous robots operate or to large industrial spaces where autonomous cleaning machines operate at night. This scenario represents well any environment that does not change significantly over time. In our simulations, the robot relies on a two-dimensional laser scanner for feature detection.

Second, we considered a real office scenario with a dynamically changing environment, people moving in it, and a wide variety of objects populating the different rooms. The experiments were carried out using visual markers that can be placed in multiple fixed locations. For this second scenario, real experiments were carried out in a real office environment with people and a variety of furniture across multiple rooms. Because of the large amount of desks, chairs, and other equipment, detecting geometric features from the environment would create multiple situations in which features cannot be detected due to either objects or people blocking the field of view of the sensors. To tackle this issue, we have utilized QR codes as markers to encode the landmark positional information. 

\subsection{Simulation Environment}

The proposed encoding approach has been implemented within the Robot Operating System (ROS) in Python. ROS is the current de facto standard for production-ready robot development~\cite{quigley2009ros, koubaa2017robot}. The simulations were carried out within the ROS/Stage environment. A TurtleBot 3 is simulated with a 2D lidar and wheel odometry. The robot was set to explore an indoor environment with a floorplan illustrated in Fig.~9. The environment is $40\times40\,m^2$, and the robot has a circular shape with a diameter of $0.35\,m$. The simulated environment contains 9 rooms with a single entrance and 6 more spaces with corridors between them. The starting exploration position of the robots is near the main door, in the bottom-left. The 2D lidar had a field of view of 270\textdegree~and produces 1080 samples (0.25\textdegree~resolution) in each scan, with a scan rate of 10\,Hz. In this experiment we did not study the effect that different odometry methods have in the exploratory mission. Instead, we used wheel odometry and varied its error to study the impact that the corresponding computational overhead had due to a larger number of hashes being calculated.

\begin{algorithm}[t]
    \kwCbk {
        \par
        \kwCal {
            \par
            $\textbf{F} = getF(data)$; \hfill \small{// Orientation-ordered $F$ set} \\
            $\textbf{F}_{cv} = getCv(\textbf{F}) \subseteq \textbf{F}$; \hfill \small{// Set of concave features} \\
            $\textbf{F}_{cc} = getCx(\textbf{F}) \subseteq \textbf{F}$; \hfill \small{// Set of convex  features}  \\
        }
        \kwDef {
            \par
            $\textbf{H} = []$; \hfill \small{// List of hashes}
        }
        \ForEach{$fp_i,\:fp_j\:\in\:\textbf{F}_{cv}$} {
            \If{$\Vert fp_i - fp_j \Vert < \delta_{dw}$} {
                $H.append(doorwayHash(fp_i, fp_j))$\;
            }
        }
        \ForEach{$fp_i\:\in\:\textbf{F}$} {
            \If{$ fp_{i+j}\in\textbf{F}_{cx}\:\:\forall j \in \{-1,0,1\} $} {
                $H.append(roomHash(fp_{i\text{-}1},fp_{i},fp_{i\text{+}1}))$\;
            }
        }
        \ForEach{$fp_i\:\in\:\textbf{F}_{cv}$} {
            \If{$isCorner(fp_i) \:\&\&\: notDoor(fp_i) $} {
                $H.append(cornerHash(fp_i))$\;
            }
        }
        \small{// Utilize any matching hashes to update the robot's} \\
        \small{// position with respect to the global reference frame} \\
        \If{$\exists\:h\in\:H\:\vert\:h\in\:NavGraph$} {
            $updateAbsolutePosition(H))$; \hfill \small{// Use matching hashes}
        }
    }
    \caption{Feature Extraction and Hash Calculation}
    \label{alg:features_and_hashes}
\end{algorithm}

\subsection*{Feature Extraction}
In the simulation experiments, we utilized three types of features to localize the robot and navigate the environment: doorways, concave corners and rooms. These are defined from the same set of $F$ feature points which we denote as Features of Interest $FoI = \{ fp_1, \dots, fp_F \}$, where $fp_i\in\mathbb{R}^3$. The feature extraction process is outlined in Algorithm \ref{alg:features_and_hashes}. The $NavGraph$ variable stores a list of hashed positions as well as an adjacency matrix with the edge hashes. A sample of this is shown in  Fig.~\ref{fig:sample_map}, subfigures (c) and (d). The function $search()$ calculates a certain number of hashes over a predefined area around the identified feature until it either finds a matching hash from $NavGraph$ or ends the search unsuccessfully. This function ensures that the hashes are reproducible even if odometry error accumulates over the inter-landmark navigation. The search area is defined based on the expected odometry error as well as the granularity of the grid utilized to define the hashes. Finally, the function $updateAbsolutePosition()$ takes the matching hashes as arguments, calculates the relative position of the robot with respect to the landmarks that have been identified and utilizes the known position of the landmarks (which is encoded in the hashes) to recalculate its own position and restart the odometry estimation.

\vspace{0.12cm}

\subsubsection*{Doorways} 
    We define doorways as any set of two concave feature points that are within two predefined distances $(\delta_{dw,min}, \delta_{dw,max})$ from each other. In our simulations and experiments, we set these distances to $\delta_{dw,min}=1.2\,m, \delta_{dw,max}=2.5\,m$. Note that these feature points might not be consecutive if we consider the ordered set of feature points by orientation.
    We define the corresponding waypoint to be encoded according to~\eqref{eq:door}:
    \begin{equation}
        H_{dw}(fp_i, fp_j)= H\left(\text{"doorway"},\frac{fp_{i}+fp_{j}}{2}, \angle fp_{i}fp_{j} \right)
        \label{eq:door}
    \end{equation}

\vspace{0.12cm}

\subsubsection*{Corners} 
    For each concave corner not in a doorway, we define its corresponding hash with~\eqref{eq:corner}:
    \begin{equation}
        H_{cv}(fp_i)= H\left(\text{"corner"}, fp_{i}, \angle fp_{i}\right)
        \label{eq:corner}
    \end{equation}
    where $\angle$ now represents the orientation of the normal vector to the wall surface at the position of the corner.

\vspace{0.12cm}

\subsubsection*{Rooms} 
    A room waypoint is defined as the centroid of any three consecutive convex points, calculated as the arithmetic mean of their positions. 
    To reduce the probability of having a mismatch in rectangular rooms where two consecutive subsets of three convex corners are visible by a robot, we add the area $\Delta$ of the triangle that the points define: 
    \begin{multline}
        H_{dw}(fp_i, fp_{i+1}, fp_{i+2})= \\
        H\left(\text{"room"},\dfrac{fp_{i}+fp_{i+1}+fp_{i+2}}{3}, \Delta_{i,i+1,i+2} \right)
    \end{multline}

\subsection{Real-Robot Experimental Settings}
\begin{figure}
    \centering
    \includegraphics[width=.48\textwidth]{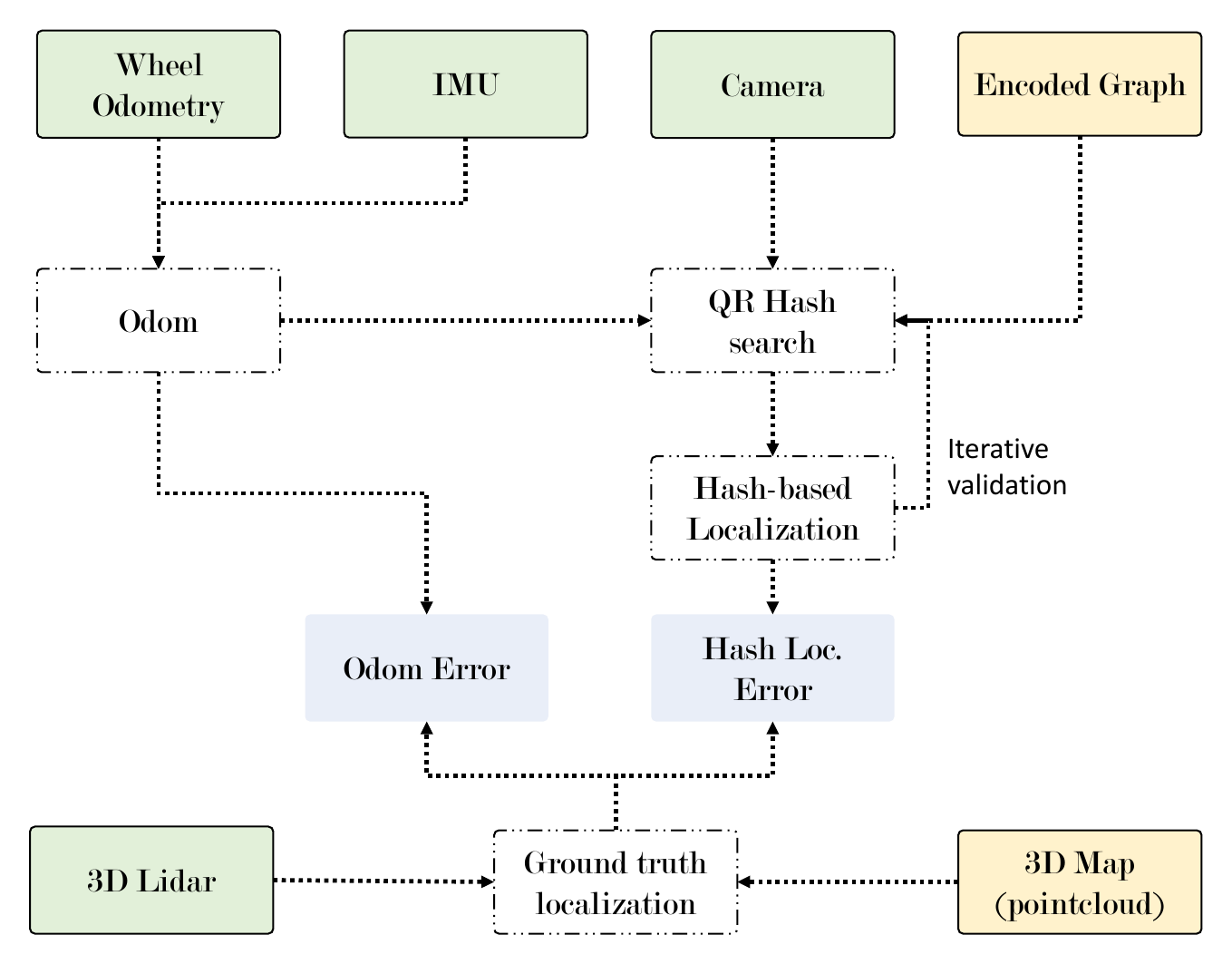}
    \caption{Data flow in the experiments. Each box represents a ROS Node which has been implemented either in C++ or Python. The outputs are the ground truth, odometry (odom) error and has-based localization error (loc. error).}
    \label{fig:ros_nodes}
\end{figure}

The experimental environment has a size of 30\,m by 25\,m. For the experiments, an EAIBOT DashGo D1 was used. We installed on the mobile robot a 16-Channel Leishen 3D Lidar, an SC-AHRS-100D2 IMU, and a Logitech c270 USB camera. The DashGo platform also provides wheel odometry from its differential drive system. The 3D lidar was used to accurately localize the landmarks and provide ground truth with map-based localization algorithms for three-dimensional point clouds introduced in~\cite{qingqing2019jdd}. The camera was used to detect the QR codes and extract the encoded information in them. Figure~\ref{fig:ros_nodes} show the implementation diagram with different ROS nodes. The 3D lidar odometry and mapping are adapted from the LeGo-LOAM-BOR package~\cite{legoloam2018}. The QR code decoding node has been written in Python using OpenCV and the Zbar library. The hash based localization node utilizes the QR codes for localization when available and the wheel and inertial odometry as an estimation between landmarks. The QR codes utilized during the experiment are of known size (12\,cm by 12\,cm), and the localization node was calibrated to map the size in pixels of a detected QR code in the camera to the distance to it. The localization also takes into account the relative orientation of the QR code.

\subsection{Feature Hashing}
We utilized SHA3-256 for hashing~\cite{czajkowski2017quantum}, which generates 32~byte hashes. It took an average of under 500~ns on an Intel(R) Core(TM) i5-6200U CPU with the pysha3 implementation in Python. If additional security is required against offline attacks on exposed hashes, other hashing algorithms such as Bcrypt~\cite{sriramya2015providing} can be utilized. Bcrypt needs around 300~ms to generate a hash with the same CPU. However, there is a trade-off between security and real-time operation as robots need to calculate multiple hashes per lidar scan. We believe that, in most applications, SHA3-256 is enough and can be utilized even in resource-constrained devices.

\begin{figure*}
    \centering
    \begin{minipage}{0.49\textwidth}
        \centering
        \setlength\figureheight{0.6\textwidth}
        \setlength\figurewidth{\textwidth}
        \footnotesize{\input{fig/sm_paths_TEST121212.tex}} \\
        \small{(a) Ground truth and estimated path during simulation.} \\
        \centering
        \setlength\figureheight{0.6\textwidth}
        \setlength\figurewidth{\textwidth}
        \footnotesize{
\begin{tikzpicture}


\definecolor{color0}{rgb}{1,0.752941176470588,0.796078431372549}
\definecolor{color1}{rgb}{0.67843137254902,0.847058823529412,0.901960784313726}

\begin{axis}[
    height=\figureheight,
    width=\figurewidth,
    axis line style={white},
    legend style={fill opacity=0.8, draw opacity=1, text opacity=1, draw=white!80!black},
    tick align=outside,
    tick pos=left,
    x grid style={white!69.0196078431373!black},
    xtick style={color=black},
    y grid style={white!90!black},
    ymajorgrids,
    ytick style={color=black},
    xmin=-2, xmax=8,
    xtick={0,2,4,6},
    xticklabels={(0.03,0),(0,0.05),(0.05,0),(0.075,0)},
    ymin=-0.144076441006528, ymax=3.02560526113709,
    ylabel={Localization Error (m)},
    xlabel={Std. of odometry noise $(\sigma_t,\sigma_r)$ in meters (translational, rotational).},
]
\addplot [black, fill=color0]
table {%
-0.7 0.415075106759834
-0.1 0.415075106759834
-0.1 0.873454159910703
-0.7 0.873454159910703
-0.7 0.415075106759834
};
\addplot [black, forget plot]
table {%
-0.4 0.415075106759834
-0.4 0
};
\addplot [black, forget plot]
table {%
-0.4 0.873454159910703
-0.4 1.16501430323376
};
\addplot [black, forget plot]
table {%
-0.55 0
-0.25 0
};
\addplot [black, forget plot]
table {%
-0.55 1.16501430323376
-0.25 1.16501430323376
};
\addplot [black, fill=color0]
table {%
1.3 0.427105229040702
1.9 0.427105229040702
1.9 0.89091473989197
1.3 0.89091473989197
1.3 0.427105229040702
};
\addplot [black, forget plot]
table {%
1.6 0.427105229040702
1.6 0
};
\addplot [black, forget plot]
table {%
1.6 0.89091473989197
1.6 1.1704129519111
};
\addplot [black, forget plot]
table {%
1.45 0
1.75 0
};
\addplot [black, forget plot]
table {%
1.45 1.1704129519111
1.75 1.1704129519111
};
\addplot [black, fill=color1]
table {%
3.3 0.708650091377608
3.9 0.708650091377608
3.9 1.47304692973965
3.3 1.47304692973965
3.3 0.708650091377608
};
\addplot [black, forget plot]
table {%
3.6 0.708650091377608
3.6 0
};
\addplot [black, forget plot]
table {%
3.6 1.47304692973965
3.6 1.92720499815684
};
\addplot [black, forget plot]
table {%
3.45 0
3.75 0
};
\addplot [black, forget plot]
table {%
3.45 1.92720499815684
3.75 1.92720499815684
};
\addplot [black, fill=color0]
table {%
5.3 1.0714265384572
5.9 1.0714265384572
5.9 2.22687546271043
5.3 2.22687546271043
5.3 1.0714265384572
};
\addplot [black, forget plot]
table {%
5.6 1.0714265384572
5.6 0
};
\addplot [black, forget plot]
table {%
5.6 2.22687546271043
5.6 2.88152882013056
};
\addplot [black, forget plot]
table {%
5.45 0
5.75 0
};
\addplot [black, forget plot]
table {%
5.45 2.88152882013056
5.75 2.88152882013056
};
\addplot [black, fill=color1]
table {%
0.1 0.178723043083498
0.7 0.178723043083498
0.7 0.413980434780265
0.1 0.413980434780265
0.1 0.178723043083498
};
\addplot [black, forget plot]
table {%
0.4 0.178723043083498
0.4 0
};
\addplot [black, forget plot]
table {%
0.4 0.413980434780265
0.4 0.647568869386604
};
\addplot [black, forget plot]
table {%
0.25 0
0.55 0
};
\addplot [black, forget plot]
table {%
0.25 0.647568869386604
0.55 0.647568869386604
};
\addplot [black, fill=color1]
table {%
2.1 0.196130333637602
2.7 0.196130333637602
2.7 0.516654820344566
2.1 0.516654820344566
2.1 0.196130333637602
};
\addplot [black, forget plot]
table {%
2.4 0.196130333637602
2.4 0
};
\addplot [black, forget plot]
table {%
2.4 0.516654820344566
2.4 0.720428044730183
};
\addplot [black, forget plot]
table {%
2.25 0
2.55 0
};
\addplot [black, forget plot]
table {%
2.25 0.720428044730183
2.55 0.720428044730183
};
\addplot [black, fill=color1]
table {%
4.1 0.244867204346871
4.7 0.244867204346871
4.7 0.588006055842012
4.1 0.588006055842012
4.1 0.244867204346871
};
\addplot [black, forget plot]
table {%
4.4 0.244867204346871
4.4 0
};
\addplot [black, forget plot]
table {%
4.4 0.588006055842012
4.4 0.865976245943324
};
\addplot [black, forget plot]
table {%
4.25 0
4.55 0
};
\addplot [black, forget plot]
table {%
4.25 0.865976245943324
4.55 0.865976245943324
};
\addplot [black, fill=color1]
table {%
6.1 0.959487966845732
6.7 0.959487966845732
6.7 2.13033045384668
6.1 2.13033045384668
6.1 0.959487966845732
};
\addplot [black, forget plot]
table {%
6.4 0.959487966845732
6.4 0
};
\addplot [black, forget plot]
table {%
6.4 2.13033045384668
6.4 2.78318883685028
};
\addplot [black, forget plot]
table {%
6.25 0
6.55 0
};
\addplot [black, forget plot]
table {%
6.25 2.78318883685028
6.55 2.78318883685028
};
\addplot [black, forget plot]
table {%
-0.7 0.631772569001384
-0.1 0.631772569001384
};
\addplot [black, forget plot]
table {%
1.3 0.643795160378899
1.9 0.643795160378899
};
\addplot [black, forget plot]
table {%
3.3 1.07090326093286
3.9 1.07090326093286
};
\addplot [black, forget plot]
table {%
5.3 1.62111326981137
5.9 1.62111326981137
};
\addplot [black, forget plot]
table {%
0.1 0.283024587321938
0.7 0.283024587321938
};
\addplot [black, forget plot]
table {%
2.1 0.318581091758525
2.7 0.318581091758525
};
\addplot [black, forget plot]
table {%
4.1 0.406165397406774
4.7 0.406165397406774
};
\addplot [black, forget plot]
table {%
6.1 1.52008370659298
6.7 1.52008370659298
};
\end{axis}

\end{tikzpicture}} \\
        \small{(b) Odometry and hash-based localization error for different odometry noise levels.}
    \end{minipage}
    \begin{minipage}{0.49\textwidth}
        \centering
        \setlength\figureheight{1.2\textwidth}
        \setlength\figurewidth{\textwidth}
        \footnotesize{
\begin{tikzpicture}


\definecolor{color0}{rgb}{1,0.752941176470588,0.796078431372549}
\definecolor{color1}{rgb}{0.67843137254902,0.847058823529412,0.901960784313726}

\begin{axis}[
    height=\figureheight,
    width=\figurewidth,
    axis line style={white},
    legend style={fill opacity=0.8, draw opacity=1, text opacity=1, draw=white!80!black},
    tick align=outside,
    tick pos=left,
    x grid style={white!69.0196078431373!black},
    xtick style={color=black},
    y grid style={white!90!black},
    ymajorgrids,
    ytick style={color=black},
    xmin=-2, xmax=10,
    xtick={0,2,4,6,8},
    xticklabels={0.4,0.2,0.1,0.05,0.025},
    ymin=3.63669079947936e-06, ymax=0.191129777886706,
    ymode=log,
    ytick={1e-07,1e-06,1e-05,0.0001,0.001,0.01,0.1,1,10},
    yticklabels={\(\displaystyle {10^{-7}}\),\(\displaystyle {10^{-6}}\),\(\displaystyle {10^{-5}}\),\(\displaystyle {10^{-4}}\),\(\displaystyle {10^{-3}}\),\(\displaystyle {10^{-2}}\),\(\displaystyle {10^{-1}}\),\(\displaystyle {10^{0}}\),\(\displaystyle {10^{1}}\)},
    ylabel={Time (s)},
    xlabel={Grid size ($\Delta$) in meters}
]
\addplot [black, fill=color0]
table {%
-0.6 0.0562741756439
-0.2 0.0562741756439
-0.2 0.0633330345154
-0.6 0.0633330345154
-0.6 0.0562741756439
};
\addplot [black, forget plot]
table {%
-0.4 0.0562741756439
-0.4 0.0457429885864
};
\addplot [black, forget plot]
table {%
-0.4 0.0633330345154
-0.4 0.0735039710999
};
\addplot [black, forget plot]
table {%
-0.5 0.0457429885864
-0.3 0.0457429885864
};
\addplot [black, forget plot]
table {%
-0.5 0.0735039710999
-0.3 0.0735039710999
};
\addplot [black, fill=color0]
table {%
1.4 0.0570418834686
1.8 0.0570418834686
1.8 0.0634760856628
1.4 0.0634760856628
1.4 0.0570418834686
};
\addplot [black, forget plot]
table {%
1.6 0.0570418834686
1.6 0.047611951828
};
\addplot [black, forget plot]
table {%
1.6 0.0634760856628
1.6 0.0729079246521
};
\addplot [black, forget plot]
table {%
1.5 0.047611951828
1.7 0.047611951828
};
\addplot [black, forget plot]
table {%
1.5 0.0729079246521
1.7 0.0729079246521
};
\addplot [black, fill=color0]
table {%
3.4 0.0576448440552
3.8 0.0576448440552
3.8 0.0632441043854
3.4 0.0632441043854
3.4 0.0576448440552
};
\addplot [black, forget plot]
table {%
3.6 0.0576448440552
3.6 0.0497159957886
};
\addplot [black, forget plot]
table {%
3.6 0.0632441043854
3.6 0.0714340209961
};
\addplot [black, forget plot]
table {%
3.5 0.0497159957886
3.7 0.0497159957886
};
\addplot [black, forget plot]
table {%
3.5 0.0714340209961
3.7 0.0714340209961
};
\addplot [black, fill=color0]
table {%
5.4 0.0567560195923
5.8 0.0567560195923
5.8 0.063218832016
5.4 0.063218832016
5.4 0.0567560195923
};
\addplot [black, forget plot]
table {%
5.6 0.0567560195923
5.6 0.0472910404205
};
\addplot [black, forget plot]
table {%
5.6 0.063218832016
5.6 0.0725090503693
};
\addplot [black, forget plot]
table {%
5.5 0.0472910404205
5.7 0.0472910404205
};
\addplot [black, forget plot]
table {%
5.5 0.0725090503693
5.7 0.0725090503693
};
\addplot [black, fill=color0]
table {%
7.4 0.0542531013489
7.8 0.0542531013489
7.8 0.0610618591309
7.4 0.0610618591309
7.4 0.0542531013489
};
\addplot [black, forget plot]
table {%
7.6 0.0542531013489
7.6 0.0441830158234
};
\addplot [black, forget plot]
table {%
7.6 0.0610618591309
7.6 0.0712380409241
};
\addplot [black, forget plot]
table {%
7.5 0.0441830158234
7.7 0.0441830158234
};
\addplot [black, forget plot]
table {%
7.5 0.0712380409241
7.7 0.0712380409241
};
\addplot [black, fill=color1]
table {%
0.2 5.96046447754e-06
0.6 5.96046447754e-06
0.6 0.000262022018433
0.2 0.000262022018433
0.2 5.96046447754e-06
};
\addplot [black, forget plot]
table {%
0.4 5.96046447754e-06
0.4 0
};
\addplot [black, forget plot]
table {%
0.4 0.000262022018433
0.4 0.000646114349365
};
\addplot [black, forget plot]
table {%
0.3 0.0000000001
0.5 0.0000000001
};
\addplot [black, forget plot]
table {%
0.3 0.000646114349365
0.5 0.000646114349365
};
\addplot [black, fill=color1]
table {%
2.2 6.19888305664e-06
2.6 6.19888305664e-06
2.6 0.000430107116699
2.2 0.000430107116699
2.2 6.19888305664e-06
};
\addplot [black, forget plot]
table {%
2.4 6.19888305664e-06
2.4 0
};
\addplot [black, forget plot]
table {%
2.4 0.000430107116699
2.4 0.001051902771
};
\addplot [black, forget plot]
table {%
2.3 0.0000000001
2.5 0.0000000001
};
\addplot [black, forget plot]
table {%
2.3 0.001051902771
2.5 0.001051902771
};
\addplot [black, fill=color1]
table {%
4.2 6.19888305664e-06
4.6 6.19888305664e-06
4.6 0.00116896629333
4.2 0.00116896629333
4.2 6.19888305664e-06
};
\addplot [black, forget plot]
table {%
4.4 6.19888305664e-06
4.4 0
};
\addplot [black, forget plot]
table {%
4.4 0.00116896629333
4.4 0.00277996063232
};
\addplot [black, forget plot]
table {%
4.3 0.0000000001
4.5 0.0000000001
};
\addplot [black, forget plot]
table {%
4.3 0.00277996063232
4.5 0.00277996063232
};
\addplot [black, fill=color1]
table {%
6.2 5.96046447754e-06
6.6 5.96046447754e-06
6.6 0.00568103790283
6.2 0.00568103790283
6.2 5.96046447754e-06
};
\addplot [black, forget plot]
table {%
6.4 5.96046447754e-06
6.4 0
};
\addplot [black, forget plot]
table {%
6.4 0.00568103790283
6.4 0.0139651298523
};
\addplot [black, forget plot]
table {%
6.3 0.0000000001
6.5 0.0000000001
};
\addplot [black, forget plot]
table {%
6.3 0.0139651298523
6.5 0.0139651298523
};
\addplot [black, fill=color1]
table {%
8.2 1.59740447998e-05
8.6 1.59740447998e-05
8.6 0.0466618537903
8.2 0.0466618537903
8.2 1.59740447998e-05
};
\addplot [black, forget plot]
table {%
8.4 1.59740447998e-05
8.4 0
};
\addplot [black, forget plot]
table {%
8.4 0.0466618537903
8.4 0.116615056992
};
\addplot [black, forget plot]
table {%
8.3 0.0000000001
8.5 0.0000000001
};
\addplot [black, forget plot]
table {%
8.3 0.116615056992
8.5 0.116615056992
};
\addplot [black, forget plot]
table {%
-0.6 0.06001496315
-0.2 0.06001496315
};
\addplot [black, forget plot]
table {%
1.4 0.0600459575653
1.8 0.0600459575653
};
\addplot [black, forget plot]
table {%
3.4 0.060683965683
3.8 0.060683965683
};
\addplot [black, forget plot]
table {%
5.4 0.0600781440735
5.8 0.0600781440735
};
\addplot [black, forget plot]
table {%
7.4 0.0576419830322
7.8 0.0576419830322
};
\addplot [black, forget plot]
table {%
0.2 8.32080841064e-05
0.6 8.32080841064e-05
};
\addplot [black, forget plot]
table {%
2.2 9.20295715332e-05
2.6 9.20295715332e-05
};
\addplot [black, forget plot]
table {%
4.2 0.000105142593384
4.6 0.000105142593384
};
\addplot [black, forget plot]
table {%
6.2 0.000108957290649
6.6 0.000108957290649
};
\addplot [black, forget plot]
table {%
8.2 0.00115609169006
8.6 0.00115609169006
};
\end{axis}

\end{tikzpicture}} \\
        \small{(c) Execution time of the different processes (feature extraction in red, hash search and matching in blue) for a varying grid size.}
    \end{minipage}
    \caption{Simulation results. Subfigure (a) shows the reconstructed path with ground truth (GT) wheel and inertial odometry (O), only doorway hashes (D), and all features: doors(D), rooms (R) and concave corners (CC). Subfigure (b) shows the odometry and hash-based localization errors for different odometry noise levels. Finally, (c) shows the execution time distribution for the feature extraction (red) and hash matching (blue) processes, where the grid size represents the search space when trying to find a hash match.}
    \label{fig:simulation_results}
\end{figure*}

\begin{figure}
    \centering
    \includegraphics[width=0.36\textwidth,height=0.24\textwidth]{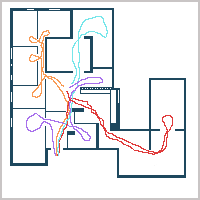}
    \caption{Illustration of the paths followed by four robots during the multi-robot collaborative exploration simulation.}
    \label{fig:cooperation}
\end{figure}

\begin{table}
    \centering
    \footnotesize
    \caption{Environment knowledge distribution during the collaborative exploration simulation.}
    \begin{tabular}{@{}lrr@{}}
        \toprule
        & \hspace{0.42cm}\textbf{Hashes found} & \hspace{0.42cm}\textbf{Area} \\
        \midrule
        \textbf{Robot$_1$} (purple)   & 12/32     & 23\% \\
        \textbf{Robot$_2$} (orange) & 16/32     & 29\% \\
        \textbf{Robot$_3$} (red)  & 15/32     & 41\% \\
        \textbf{Robot$_4$} (cian)   & 11/32     & 27\% \\
        \bottomrule
    \end{tabular}
    \label{tab:coop}
\end{table}

\begin{figure*}
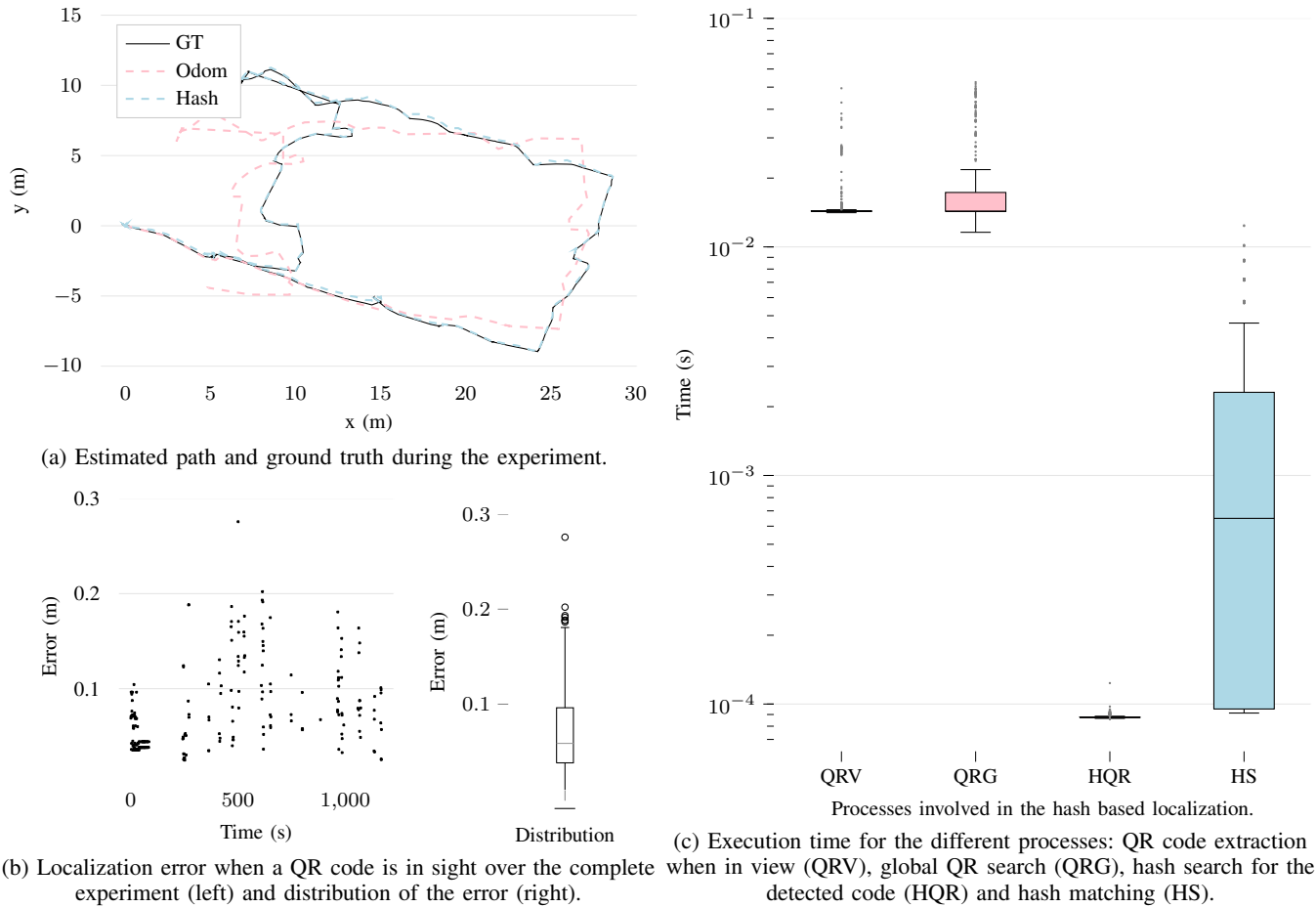


    \centering
    \begin{minipage}{0.49\textwidth}
            \centering
            \setlength\figureheight{0.72\textwidth}
            \setlength\figurewidth{\textwidth}
            \footnotesize{\input{exp/experiment_paths_xs_v2.tex}} \\
            \small{(a) Estimated path and ground truth during the experiment.}
        $ $ \\[+6pt]
            \centering
            \setlength\figureheight{0.6\textwidth}
            \setlength\figurewidth{0.6\textwidth}
            \footnotesize{\input{exp/experiment_qr_errors_xs.tex}}
            \setlength\figureheight{0.6\textwidth}
            \setlength\figurewidth{0.35\textwidth}
            \footnotesize{
\begin{tikzpicture}

\definecolor{color0}{rgb}{0.6,0.6,0.6}

\begin{axis}[
    axis line style={white},
    ylabel={Error (m)},
    xlabel={Distribution},
    height=\figureheight,
    legend cell align={left},
    legend style={at={(0.91,0.5)}, anchor=east, draw=white!80.0!black},
    log basis y={10},
    tick align=outside,
    tick pos=left,
    width=\figurewidth,
    x grid style={white!69.01960784313725!black},
    y grid style={white!69.01960784313725!black},
    xmin=0.5, xmax=1.5,
    ymin=0.01, ymax=0.3,
    xtick={1},
    xticklabels={|}
]
\addplot [black]
table {%
0.925 0.0382378239497
1.075 0.0382378239497
1.075 0.096221912142725
0.925 0.096221912142725
0.925 0.0382378239497
};
\addplot [black]
table {%
1 0.0382378239497
1 0.000833320992284
};
\addplot [black]
table {%
1 0.096221912142725
1 0.180713529514
};
\addplot [black]
table {%
0.9625 0.000833320992284
1.0375 0.000833320992284
};
\addplot [black]
table {%
0.9625 0.180713529514
1.0375 0.180713529514
};
\addplot [black, mark=o, mark size=1.23, mark options={solid,fill opacity=0}, only marks]
table {%
1 0.188296622417
1 0.188296622417
1 0.186446148731
1 0.275758925617
1 0.193223439755
1 0.20212547132
1 0.191388804662
};
\addplot [color0]
table {%
0.925 0.05873929177985
1.075 0.05873929177985
};
\end{axis}

\end{tikzpicture}} \\
            \small{(b) Localization error when a QR code is in sight over the complete experiment (left) and distribution of the error (right).}
    \end{minipage}
    \begin{minipage}{0.49\textwidth}
        \centering
        \setlength\figureheight{1.3\textwidth}
        \setlength\figurewidth{\textwidth}
        \footnotesize{\input{exp/times.tex}} \\
        \small{(c) Execution time for the different processes: QR code extraction when in view (QRV), global QR search (QRG), hash search for the detected code (HQR) and hash matching (HS).}
    \end{minipage}
    
    \caption{Experiment results. Subfigure (a) shows the reconstructed path with ground truth (GT), wheel and inertial odometry (Odom) and hash-based localization (Hash). In (b), we show a scatter plot with the localization error every time that a QR code is within the field of view of the camera, and a boxplot with the global error distribution. Finally, (c) shows the execution time for the different processes involved in the localization, with the hash search and matching being one to two orders of magnitude below the image processing processes that would still be in use in a traditional approach.}
    \label{fig:experiment_results}
    
\end{figure*}

\section{Simulation and Experimental Results}
\label{sec:Results}

We carried out a series of simulations with one and multiple robots to evaluate mainly the cost of utilizing hash matching for localization and navigation, but also the impact on accuracy of the encoded landmarks.

\subsection{Metrics}
To evaluate the simulation results, we measured the absolute localization error of the robot with odometry only and hash matching. Furthermore, we analyzed the distribution of the computational load among the different tasks that the robots are carrying out: feature extraction, hash calculation and hash matching. In the simulations, we also measured the effect of the odometry-based localization noise and the choice of spatial granularity for landmark positions.

\subsection{Single-Robot Simulation Results}
The aim of the simulations is to prove whether our encoded landmark localization and navigation scheme adds a significant computational overhead or not.

Fig.~\ref{fig:simulation_results}\,(a) shows the path recovered from odometry measurements and hash-based localization with doorway hashes only, and all three types of hashes, together with the ground truth (GT). The data was recorded over 150\,s; the translation odometry noise was set with $\sigma_t=0.03$, and rotation noise with $\sigma_r=0.05$. The error distribution for the two methods is given in Fig.~\ref{fig:simulation_results}\,(c). 

In the simulated environment, we predefined the position of landmarks with an accuracy of $0.1$\,m. Therefore, when analyzing the errors in Fig.~\ref{fig:simulation_results}\,(b), any values below $0.15$~m represent virtually zero error. Fig.~\ref{fig:simulation_results}\,(b) shows that the localization method is robust even when the odometry error increases significantly ($\sigma_t=0.05$). However, there is a limit, around $\sigma_t=0.06$, for which the size of the environment is big enough so that the robot is unable to match the correspondent landmark hashes due to accumulated odometry drift. In order to calculate these hashes, we assumed an error tolerance with respect to its estimated position of $\pm0.5$\,m, independently of the size of the grid utilized to locate the landmarks and generate the hashes. This limit defines the computational time required for the hash search together with the grid size.

Regarding the computational overhead necessary to calculate the hashes, estimate the robot's position, and perform path planning accordingly, Fig.~\ref{fig:simulation_results}\,(c) shows the distribution of computational time utilized to extract the set of features, or points of interest, from the raw lidar data and the distribution of computational time utilized in calculating and matching hashes. For an error tolerance of $\pm0.5$\,m, the graphic shows situations in which the robot tests up to 9, 25, 121, 441 and 1681 grid positions, respectively. The search for a hash match is gradually done in a spiral manner around the estimated position and within the aforementioned error tolerance. These results show that even with fine-grained grid search, in average the time required to localize the robot based on hashes is two orders of magnitude smaller than the time required to extract features from lidar data. In the worst-case scenario, the time required can be comparable, with an equivalent order of magnitude for both hash matching and feature extraction. 

\subsection{Multi-robot Exploration Simulation Results}
In terms of cooperative exploration, we provide a qualitative analysis of a four-robot cooperation. Fig.~\ref{fig:cooperation} shows the paths of four robots exploring different areas of the same simulation environment. By utilizing encoded landmarks, these robots can share their progress upon meeting in the center of the maze without revealing the raw data they have acquired. If the mission's nature is to perform surveillance or detect a series of items, robots do not need to store raw map data at all. Nonetheless, even if they did, the knowledge of the objective environment remains divided, as shown in Table~\ref{tab:coop}. In this case, robots acquire in average raw data form only 30\% of the objective exploration environment, and 41\% at most.

\subsection{Experimental Results}
Fig.~\ref{fig:experiment_results}\,(a) shows the path recovered from odometry measurements and hash-based localization (QR codes). The error in the odometry is significantly higher than in the simulation experiments due to a drift in the yaw measurements. However, the translational odometry error is much smaller. The hash-based localization is able to correct this orientation whenever a QR code is within the field of view, and therefore it does not suffer from the yaw drift. The maximum hash-based localization error that we observed was of 41.3\,cm (between observations of landmarks and owing to the accumulated odometry drift). This allowed the utilization of a fine grid of 2\,cm for calculating the landmark hashes. We set, experimentally, a $1\,m^2$ hash search area around the estimated location.

A total of 23\,QR codes were installed in the office environment, and the tests were done with a small number of persons in their offices. Out of those, 17\,QR codes were utilized by the robot during its navigation.The mission times at which at least one code was in sight, the error from each observation, and the global error distribution are shown in Fig.~\ref{fig:experiment_results}\,(b). The execution time of the hash matching algorithm was on average over one order of magnitude smaller than the time required to extract the QR codes from camera images. Thus, the overhead was mostly negligible. Only in a reduced number of occasions was the latency of these two processes comparable, as Fig.~\ref{fig:experiment_results}\,(c) shows.

\subsection{Viability and Usability}

We have seen that the computational overhead added when encoding landmarks is mostly negligible. Thus, our approach could be incorporated on top of many existing navigation and localization schemes, whether they are landmark-based or not, to increase the level of security if the error tolerance allows. 

This approach has additional uses when more than one robot is taken into consideration. In multi-robot cooperation, different robots can share their plans, progress or position (based on the navigation graph only) with others by utilizing the same hashes or parts of them. This would reduce the possibility of raw data being exposed but also virtually eliminate the options for attackers or byzantine agents to affect the mission, as has been shown in~\cite{ferrer2019secure}. Moreover, as we have shown with the multi-robot simulations, the framework allows for multi-robot exploration or other collaborative mission, reducing the fraction of raw data or environment information that has to be made available to each individual unit.

\subsection{Validation of sensor data}
In the navigation experiments conducted, we were also able to leverage the encoded landmarks for validating the sensor data leading to odometry estimations. In these experiments, the robot was able to estimate the error (i.e., drift) in the odometry as a function of the computational time required to decode the landmark position (i.e., the number of hashes that need to be tested against the encoded map information). If the odometry error is too large, or a sensor malfunctions, then the hash cannot be decoded unless the hash search radius and computational timeout are extended consequently. In both the simulations and real-robot experiments, we utilized only IMU and wheel odometry data (e.g., versus visual-inertial or lidar odometry) to evaluate the performance of the proposed methods in more adversarial conditions where the odometry error between landmarks may increase significantly.

\subsection{Trade-offs and security considerations}
We have shown through simulation and real-world experiments that the proposed framework has negligible impact in terms of usage of computational resources. In the simulations, performing a hash search in a grid map with a resolution of 2.5\,cm requires in average over an order of magnitude less computational resources than the lidar feature extraction that is inherent to any localization process. Nonetheless, for each specific application in which an encoded instruction graph is used, there will be a certain size of the hash search space where the instruction decoding is no longer negligible.

Another trade-off occurs between real-time operation and security. The larger the hash search space, the more computational resources are needed to perform a brute-force attack on the encoded information. Nonetheless, performing such attack requires information on the encoding algorithm, therefore requiring physical access to the robot or access to the control algorithms and data processing stack. The specific threshold on the hash search granularity will be defined, among other factors, by the hashing algorithm, and the ways in which the hashes are generated (e.g., involving time or unknown external inputs from the controller can, in many situations, render the brute-force attacks unfeasible).

In terms of the resilience of the proposed framework against adversarial attacks, the main security vulnerability that we have detected is the ability of an attacker to reproduce the encoded commands even without decoding them, potentially triggering the robot into repeating actions. If data is spoofed when transmitted to the robot, the robot's behavior could be studied under different encoded commands, and then an attacker could trigger a known mission. While this cannot be completely mitigated within our proposed approach, we have introduced time-checked actions and event-triggered actions. If a one-time action is required and either the start of the mission or its timing is known, then it is feasible to include the time component into the encoded instructions to avoid repeated actions even if the encoded data is spoofed. Moreover, other generic strategies designed against data spoofing could be introduced on top of our framework.

\section{Conclusion and Future Work}

Security and safety in robotics are crucial aspects to consider given the current surge of autonomous systems.  In this direction, further research is needed on the validation of data at the different layers of robotic systems, and in particular the validation of the interaction of a robot with its environment. This interaction often starts with navigation, which has been studied in this paper.

Navigation and localization in autonomous robots require large amounts of raw data for long-term operation, either given {\it a priori} by a mission controller or acquired by robots while performing their missions. In addition, validating the integrity of both mission instructions and sensor data without any external feedback is an open problem. In this paper, we have presented a framework that enables robots to validate both the correct operation of their onboard hardware and sensors, and the integrity of information received from an external controller.

In particular, to the best of our knowledge, this paper introduces and evaluates the first framework which allows robots to effectively perform their missions while also performing end-to-end validation of information, demonstrated on navigation and localization with encoded landmarks. We have shown that utilizing an encoded navigation graph adds only a negligible computational overhead even when high-accuracy positioning is required.

The end-to-end validation scheme demonstrated in this work for navigation tasks can be naturally extended to cover virtually all domains of robotic operation. In future work, we will focus our research efforts on experimentation in more realistic environments, and in particular industrial settings. We will aim at extending this approach to other forms of interaction between a robot and its environment, from multi-robot collaborative assembly to human-robot interaction and control.

\section*{Acknowledgements}

This work was supported by the Academy of Finland's AutoSOS (grant number 328755) and RoboMesh (grant number 336061) projects. This project has also received funding from the European Union’s Horizon 2020 research and innovation programme under the Marie Skłodowska-Curie grant agreement No. 751615, and the Finnish Foundation for Technology Promotion under grant number 8089.

\bibliographystyle{IEEEtran}
\bibliography{main.bib}

\newpage
\begin{IEEEbiography}[{\includegraphics[width=0.94in,height=1.25in,clip,keepaspectratio]{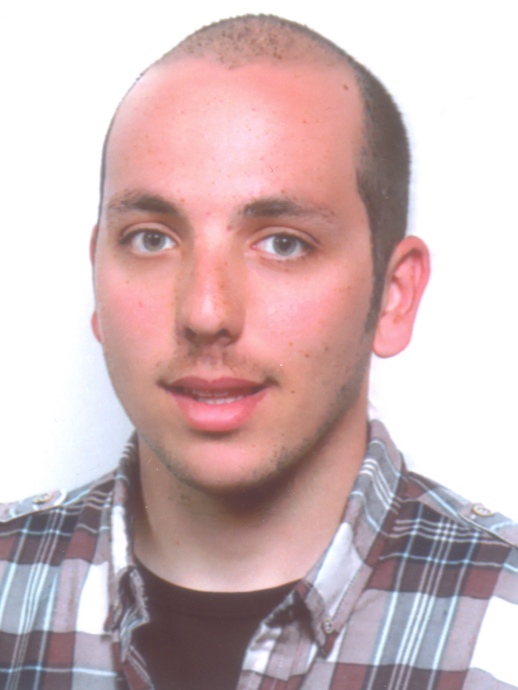}}]{Jorge Peña Queralta} received B.S. degrees in mathematics and physics engineering from UPC BarcelonaTech, Spain, in 2016, a M.Sc. (Tech.) degree in Information and Communication Science and Technology from the University of Turku, Finland, and a M. Eng. degree in Electronics and Communication Engineering from Fudan University, China, in 2018. Since 2018, he has been a researcher and doctoral candidate at the Turku Intelligent Embedded and Robotic Systems (TIERS) Group, University of Turku. His research interests include multi-robot systems, collaborative autonomy, distributed perception, and edge computing.
\end{IEEEbiography}

\vspace{-1.5em}

\begin{IEEEbiography}[{\includegraphics[width=0.94in,height=1.25in,clip,keepaspectratio]{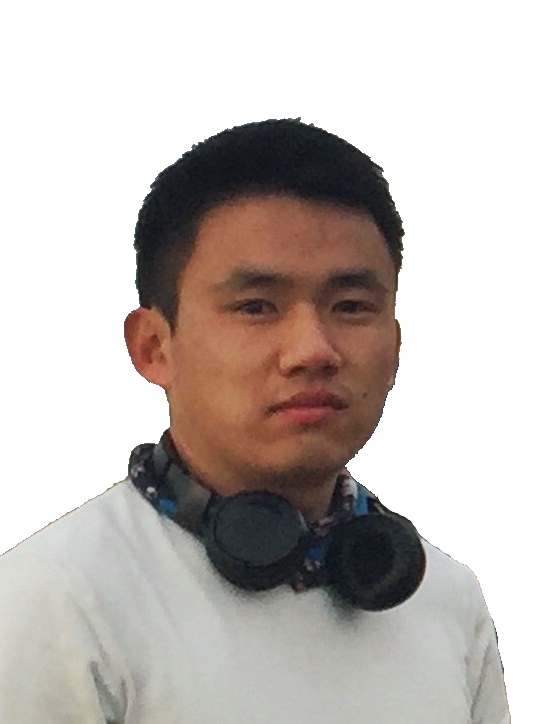}}]{Qingqing Li} received his B.S. degree in Electrical Engineering and his M.Sc. degree in Electronics and Communication Engineering, from Fudan University, Shanghai, China, in 2016 and 2018, respectively. He also holds a M.Sc. degree in Information and Communication Science and Technology from the University of Turku, Finland. Since 2019, he has been a researcher at the Turku Intelligent Embedded and Robotic Systems (TIERS) Group, University of Turku. His research interests include sensor fusion for autonomous robots and vehicles, three-dimensional point cloud data analysis and multi-robot collaboration.
\end{IEEEbiography}

\vspace{-1.5em}

\begin{IEEEbiography}[{\includegraphics[width=0.94in,height=1.25in,clip,keepaspectratio]{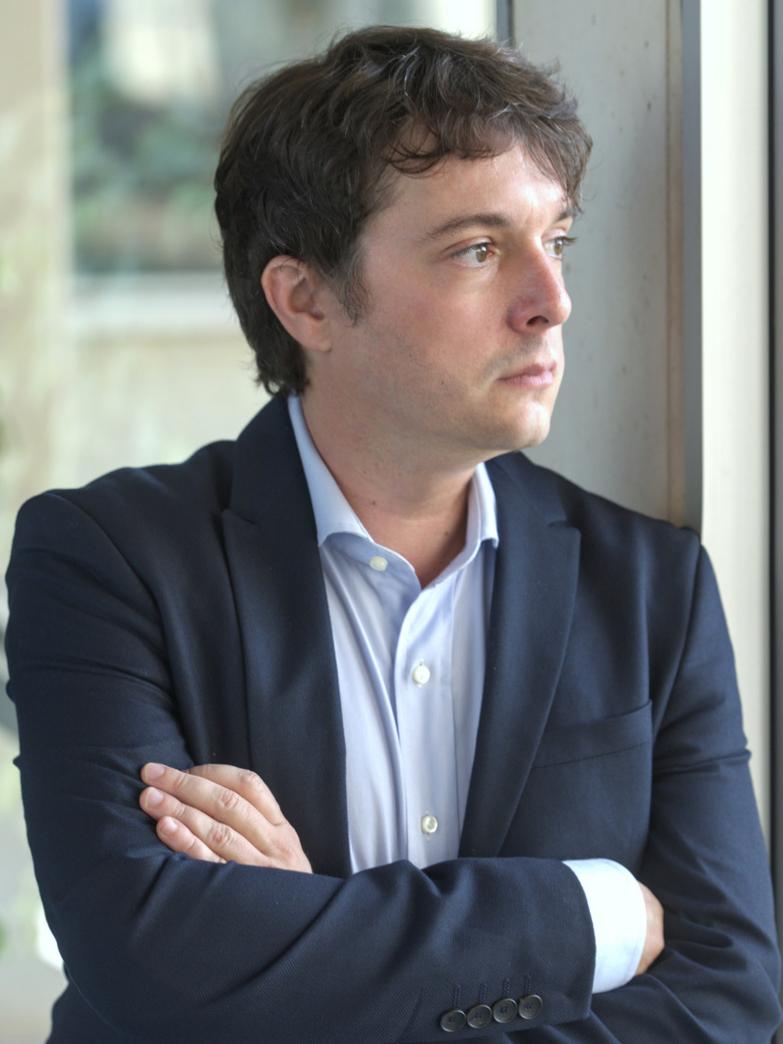}}]{Eduardo Castelló Ferrer} received his Bsc.(Hons) Intelligent Systems from University of Portsmouth (UK) and his M. Eng and PhD degrees in robotics engineering from Osaka University (Japan). His experience and interests comprise robotics, blockchain technology, and complex systems. He is a postdoctoral Marie Curie Fellow (MSCA) at the MIT Connection Science and the MIT Media Lab where he explores the combination of robotic systems and blockchain technology. His work focuses on implementing new security, behavior, and business models for distributed robotics by using novel cryptographic methods.
\end{IEEEbiography}

\vspace{-1.5em}

\begin{IEEEbiography}[{\includegraphics[width=1in,height=1.25in,clip,keepaspectratio]{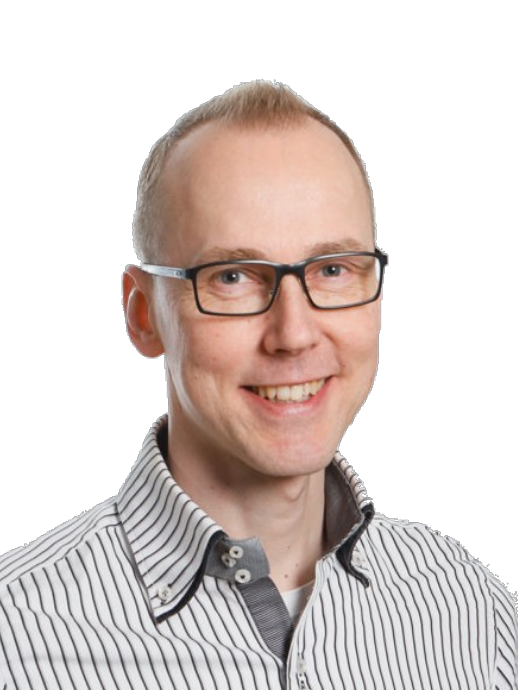}}]{Tomi Westerlund} is an Associate Professor of Autonomous Systems and Robotics at the University of Turku and a Research Professor at Wuxi Institute of Fudan University, Wuxi, China. Dr. Westerlund leads the Turku Intelligent Embedded and Robotic Systems research group (tiers.utu.fi), University of Turku, Finland. His current research interest is in the areas of Industrial IoT, smart cities and autonomous vehicles (aerial, ground and surface) as well as (co-)robots. In all these application areas, the core research interests are in multi-robot systems, collaborative sensing, interoperability, fog and edge computing, and edge AI.
\end{IEEEbiography}

\vspace*{\fill}

\end{document}